\documentclass[letterpaper]{article} % DO NOT CHANGE THIS
\usepackage{aaai2026}  % DO NOT CHANGE THIS
\usepackage{times}  % DO NOT CHANGE THIS
\usepackage{helvet}  % DO NOT CHANGE THIS
\usepackage{courier}  % DO NOT CHANGE THIS
\usepackage[hyphens]{url}  % DO NOT CHANGE THIS
\usepackage{graphicx} % DO NOT CHANGE THIS
\usepackage{natbib}  % DO NOT CHANGE THIS AND DO NOT ADD ANY OPTIONS TO IT
\usepackage{caption} % DO NOT CHANGE THIS AND DO NOT ADD ANY OPTIONS TO IT
\usepackage{algorithm}
\usepackage{algorithmic}

\usepackage{times}  % DO NOT CHANGE THIS
\usepackage{helvet}  % DO NOT CHANGE THIS
\usepackage{courier}  % DO NOT CHANGE THIS
\usepackage[hyphens]{url}  % DO NOT CHANGE THIS
\usepackage{graphicx} % DO NOT CHANGE THIS
\usepackage{natbib}  % DO NOT CHANGE THIS AND DO NOT ADD ANY OPTIONS TO IT
\usepackage{caption} % DO NOT CHANGE THIS AND DO NOT ADD ANY OPTIONS TO IT
\usepackage{algorithm}
\usepackage{algorithmic}
\usepackage[utf8]{inputenc} % allow utf-8 input
\usepackage[T1]{fontenc}    % use 8-bit T1 fonts
\usepackage{url}            % simple URL typesetting
\usepackage{booktabs}       % professional-quality tables
\usepackage{amsfonts}       % blackboard math symbols
\usepackage{nicefrac}       % compact symbols for 1/2, etc.
\usepackage{microtype}      % microtypography
\usepackage{xcolor}         % colors
\usepackage{graphicx}
\usepackage{url}
\usepackage{bm}
\usepackage{xspace}
\usepackage{multirow} 
\usepackage{amsthm}
\usepackage{amsmath}
\theoremstyle{plain}

\usepackage{float}
\theoremstyle{plain}
\newtheorem*{theorem*}{Theorem}
\usepackage{amsmath}
\theoremstyle{remark}

\usepackage{booktabs}
\usepackage{multirow}
\usepackage{cancel}
\usepackage{xcolor}

\usepackage{newfloat}
\usepackage{listings}

\usepackage{newfloat}
\usepackage{listings}
\DeclareCaptionStyle{ruled}{labelfont=normalfont,labelsep=colon,strut=off} % DO NOT CHANGE THIS
\floatstyle{ruled}
\newfloat{listing}{tb}{lst}{}
\floatname{listing}{Listing}
\newcommand{\ours}{CogGRAG}

\makeatletter
\renewcommand\@fnsymbol[1]{\ensuremath{\ifcase#1\or *\or \dagger\or \ddagger
   \or \mathsection\or \mathparagraph\or \|\or **\or \dagger\dagger
   \or \ddagger\ddagger \else\@ctrerr\fi}}
\makeatother

\title{Human Cognition Inspired RAG with Knowledge Graph for Complex Problem Solving}

\author {
    Yao Cheng\textsuperscript{\rm 1},
    Yibo Zhao\textsuperscript{\rm 1},
    Jiapeng Zhu\textsuperscript{\rm 1},
    Yao Liu\textsuperscript{\rm 1}\thanks{Corresponding author.},
    Xing Sun\textsuperscript{\rm 2},
    Xiang Li\textsuperscript{\rm 1}\footnotemark[1]
}

\affiliations{
    \textsuperscript{\rm 1}School of Data Science and Engineering, East China Normal University, China\\
    \textsuperscript{\rm 2}Tencent Youtu Lab, China\\
  
}

\usepackage{bibentry}
\begin{document}

\maketitle

\begin{abstract}
Large Language Models (LLMs) have demonstrated significant potential across various domains. However, they often struggle with integrating external knowledge and performing complex reasoning, leading to hallucinations and unreliable outputs. Retrieval Augmented Generation (RAG) has emerged as a promising paradigm to mitigate these issues by incorporating external knowledge. 
% Yet, conventional RAG approaches—especially those based on vector similarity—fail to effectively handle relational structures and multi-step reasoning. In this work, we propose CogGRAG, a human cognition-inspired, graph-based RAG framework designed for Knowledge Graph Question Answering (KGQA). 
% CogGRAG mimics human reasoning through a three-stage process: (1) top-down problem decomposition via mind map construction; (2) structured retrieval of local and global knowledge from external Knowledge Graphs (KGs); and (3) bottom-up reasoning with self-verification. Unlike previous tree-based decomposition methods such as MindMap or Graph-CoT, CogGRAG unifies the entire reasoning process under a global mind map with early-stage, graph-structured retrieval and integrates dual-process verification to mitigate error propagation. Extensive experiments demonstrate that CogGRAG outperforms existing methods in both accuracy and reliability.
Yet, conventional RAG approaches, especially those based on vector similarity, fail to effectively capture relational dependencies and support multi-step reasoning.
In this work, we propose \ours, a human cognition-inspired, graph-based RAG framework designed for Knowledge Graph Question Answering (KGQA). 
CogGRAG models the reasoning process as a tree-structured mind map that decomposes the original problem into interrelated subproblems and explicitly encodes their semantic relationships. 
This structure not only provides a global view to guide subsequent retrieval and reasoning but also enables self-consistent verification across reasoning paths. 
The framework operates in three stages: (1) top-down problem decomposition via mind map construction, (2) structured retrieval of both local and global knowledge from external Knowledge Graphs (KGs), and (3) bottom-up reasoning with dual-process self-verification.
Unlike previous tree-based decomposition methods such as MindMap or Graph-CoT, CogGRAG unifies problem decomposition, knowledge retrieval, and reasoning under a single graph-structured cognitive framework, allowing early integration of relational knowledge and adaptive verification. Extensive experiments demonstrate that CogGRAG achieves superior accuracy and reliability compared to existing methods.
% We provide our code and data here: 
% \url{https://github.com/cy623/RAG.git}.
\end{abstract}

\begin{links}
    \link{Code}{https://github.com/cy623/RAG.git}
    % \link{Datasets}{https://aaai.org/example/datasets}
    % \link{Extended version}{https://aaai.org/example/extended-version}
\end{links}

\section{Introduction}
As a foundational technology for artificial general intelligence (AGI), large language models (LLMs) have achieved remarkable success in practical applications, demonstrating transformative potential across a wide range of domains~\cite{llama2, llama3,qwen2}.
Their ability to process, generate, and reason with natural language has enabled significant advancements in areas such as machine translation~\cite{zhu2023multilingual}, text summarization~\cite{basyal2023text}, and question answering~\cite{pan2306unifying}.
Despite their impressive performance, LLMs still face significant limitations in knowledge integration beyond their pre-trained data boundaries.
% which
These 
limitations 
often lead to the generation of plausible but factually incorrect responses, a phenomenon commonly referred to as \emph{hallucinations}, which undermines the reliability of LLMs in critical applications.

\begin{figure*}[h]
  \centering  \includegraphics[width=0.8\linewidth]{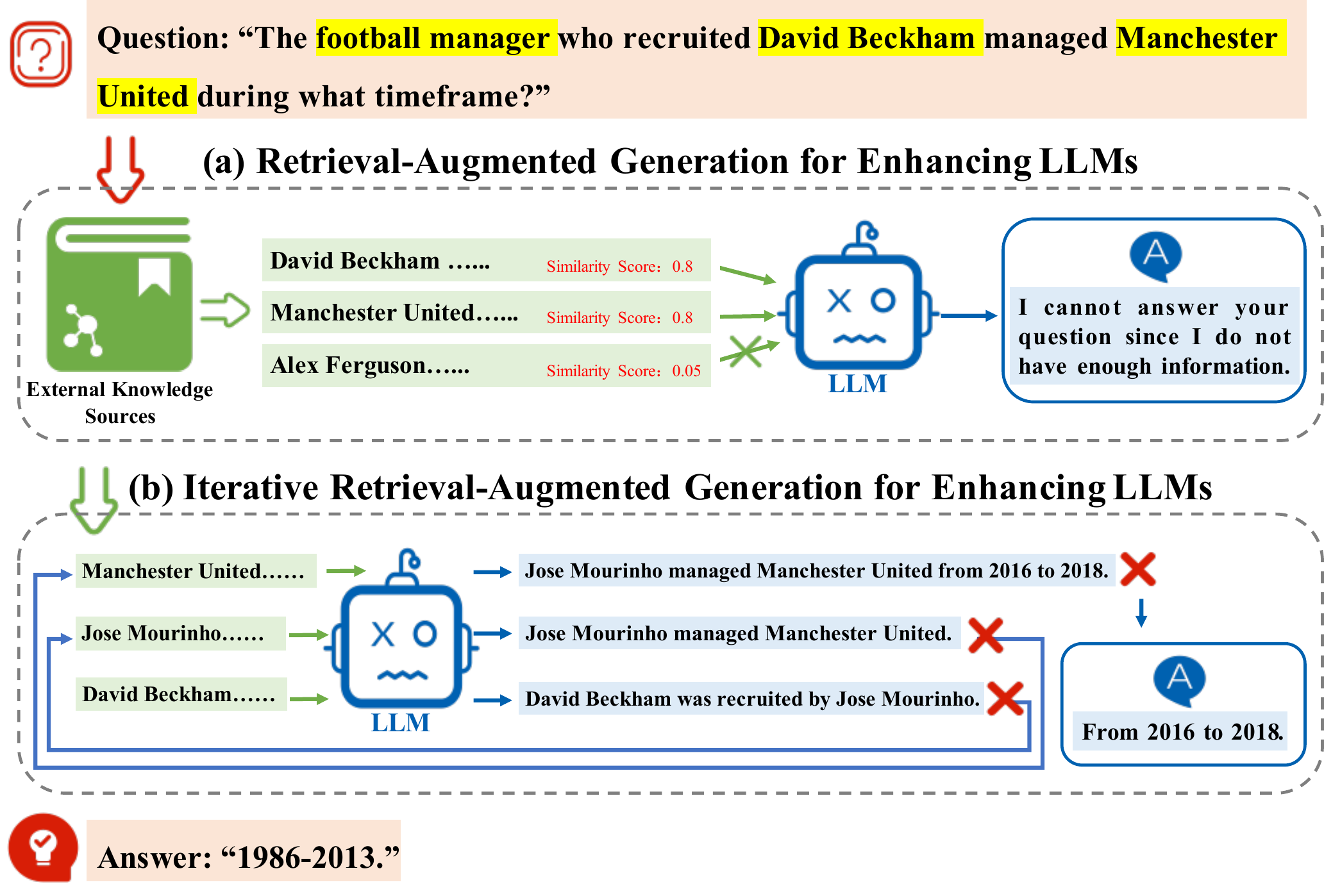}
  \caption{Representative workflow of two Retrieval-Augmented Generation paradigms for enhancing LLMs. 
  }
  \label{fg:existing}
\end{figure*}

To mitigate hallucinations, 
% in LLMs
Retrieval Augmented Generation (RAG)~\cite{niu2023ragtruth,gao2023retrieval,graphrag,xin2024atomr,chu2024beamaggr,cao2023probabilistic,wu2025pa,zhao20252graphrag} has emerged as a promising paradigm,
% to mitigate hallucinations in LLMs, 
significantly improving the accuracy and reliability of LLM-generated contents through the integration of external knowledge.
% Although RAG enables LLMs to access external knowledge sources and partially mitigates the hallucination issue, it faces significant limitations in processing and utilizing complex relational information. 
However, while RAG successfully mitigates certain aspects of hallucination,
% through external knowledge integration, 
it still exhibits inherent limitations in processing complex relational information. 
% Specifically, t
As shown in Figure~\ref{fg:existing} (a), 
the core limitation of standard RAG systems lies in their reliance on vector-based similarity matching, which processes knowledge segments as isolated units without capturing their contextual interdependencies or semantic relationships~\cite{graph_cot}. 
Consequently, traditional RAG implementations are inadequate for supporting advanced reasoning capabilities in LLMs, particularly in scenarios requiring complex problem-solving, multi-step inference, or sophisticated knowledge integration.

% Recent advancements in knowledge-enhanced LLMs have witnessed the emergence of Graph RAG methodologies, which 
Recently, 
graph-based RAG~\cite{graphrag,tog2,graph_cot,gnnrag,xiong2024interactive}
has been proposed
% aims 
to address the limitations of conventional RAG systems by incorporating deep structural information from external knowledge sources.
These approaches typically utilize KGs to model complex relation patterns within external knowledge bases, employing structured triple representations (<entity, relation, entity>) to integrate fragmented information across multiple document segments. 
% This graph RAG framework establishes a global knowledge network, enabling more comprehensive contextual understanding and facilitating advanced reasoning capabilities in LLMs.
While 
% graph retrieval augmented generation (G-RAG)
graph-based RAG has shown promising results in mitigating hallucination and improving factual accuracy, several challenges remain unresolved:

\noindent $\bullet$
\textbf{Lack of holistic reasoning structures. }
Complex problems cannot be resolved through simple queries; they typically require multi-step reasoning to derive the final answer. 
Existing methods often adopt iterative or sequential inference pipelines that are prone to cascading errors due to the absence of global reasoning plans~\cite{graph_cot,gnnrag}.
As shown in Figure~\ref{fg:existing} (b), 
each step in the iterative process relies on the result of the previous step, indicating that 
errors occurred at previous steps can propagate to subsequent steps.

\noindent $\bullet$ \textbf{Absence of verification mechanisms.}
Despite the integration of external knowledge sources, LLMs remain prone to generating inaccurate or fabricated responses when confronted with retrieval errors or insufficient knowledge coverage.
Most approaches do not incorporate explicit self-verification, making them vulnerable to retrieval errors or spurious reasoning paths.

To address these challenges, we propose \ours, a \textbf{Cog}nition inspired \textbf{G}raph \textbf{RAG} framework, designed to enhance the complex problem-solving capabilities of LLMs in Knowledge Graph Question Answering (KGQA) tasks.
\ours\ is motivated by dual-process theories in human cognition, where problem solving involves both structured decomposition and iterative reflection.
Our key contributions are as follows:

\noindent $\bullet$ \textbf{Human-inspired decomposition.}
% \ours\ introduces a top-down question decomposition strategy that constructs a structured “mind map,” allowing the system to identify and organize subproblems before retrieval. Unlike MindMap~\cite{mindmap} or Graph-CoT~\cite{graph_cot}, which rely on sequential or shallow decomposition, \ours\ builds a hierarchical and globally consistent structure guiding all downstream steps.
\ours\ introduces a top-down question decomposition strategy that constructs a tree-structured mind map, explicitly modeling semantic relationships among subproblems. This representation enables the system to capture dependencies, causal links, and reasoning order between subcomponents, providing a global structure that guides subsequent retrieval and reasoning.

\noindent $\bullet$ \textbf{Hierarchical structured retrieval.}
 \ours\ performs both local-level (entity/triple-based) and global-level (subgraph) retrieval based on the mind map. This dual-level retrieval strategy ensures precise and context-rich knowledge grounding.

\noindent $\bullet$ \textbf{Self-verifying reasoning.}
 Inspired by cognitive self-reflection, \ours\ employs a dual-LLM verification mechanism. The reasoning LLM generates answers, which are then reviewed by a separate verifier LLM to detect and revise erroneous outputs. This design reduces hallucinations and improves output faithfulness.

We validate \ours\ across three general KGQA benchmarks (HotpotQA, CWQ, WebQSP) and one domain-specific KGQA dataset (GRBench). Empirical results demonstrate that \ours\ consistently outperforms strong baselines in both accuracy and hallucination suppression, under multiple LLM backbones.

\section{Related Work}

\begin{figure*}[h]
  \centering  \includegraphics[width=0.85\linewidth]{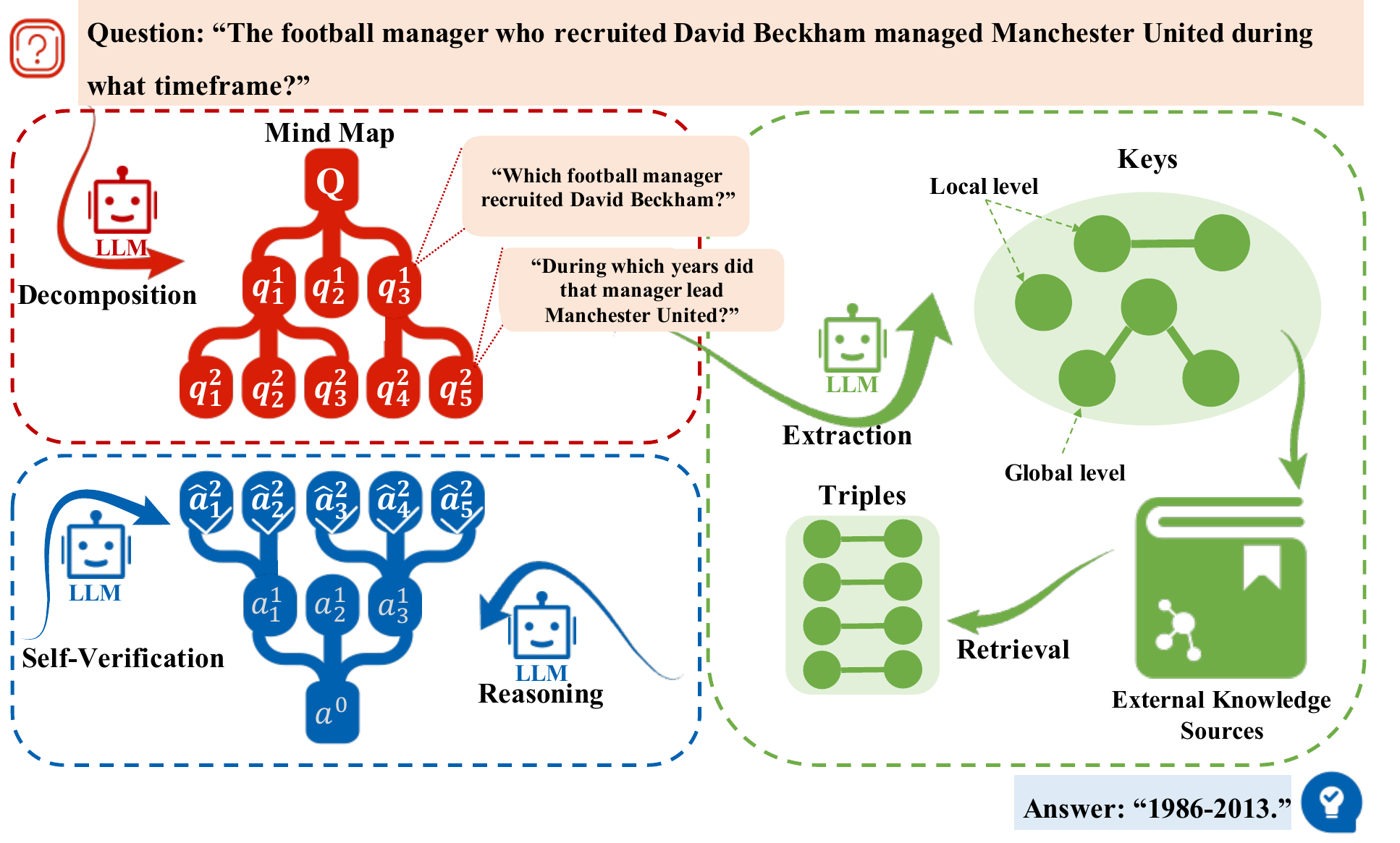}
  \caption{The overall process of \ours.
  Given a target question $Q$, \ours\ first prompts the LLM to decompose it into a hierarchy of sub-problems in a top-down manner, constructing a structured mind map. 
  % Subsequently, \ours\ prompts the LLM to extract both local and global level key information from these sub-problems. 
  Subsequently, \ours\ prompts the LLM to extract both local level (entities and triples) and global level (subgraphs) key information from these questions.
  Finally, \ours\ guides the LLM to perform bottom-up reasoning and verification based on the mind map and the retrieved knowledge, until the final answer is derived. 
  }
  \label{fg:overall}
\end{figure*}

\textbf{Reasoning with LLM Prompting.}
Recent advancements in prompt engineering have demonstrated that state-of-the-art prompting techniques can significantly enhance the reasoning capabilities of LLMs on complex problems~\cite{prompt_cot, prompt_tree, prompt_graph}.
Chain of Thought (CoT~\cite{prompt_cot} explores how generating a chain of thought—a series of intermediate reasoning steps—significantly improves the ability of large language models to perform complex reasoning.
Tree of Thoughts (ToT)~\cite{prompt_tree} introduces a new framework for language model inference that generalizes the popular Chain of Thought approach to prompting language models, enabling exploration of coherent units of text (thoughts) as intermediate steps toward problem-solving.
The Graph of Thoughts (GoT)~\cite{prompt_graph} models the information generated by a LLM as an arbitrary graph, enabling LLM reasoning to more closely resemble human thinking or brain mechanisms.
However, these methods remain constrained by the limitations of the model's pre-trained knowledge base and are unable to address hallucination issues stemming from the lack of access to external, up-to-date information.

\textbf{Knowledge Graph Augmented LLM.}
KGs~\cite{wikidata} offer distinct advantages in enhancing LLMs with structured external knowledge. 
Early graph-based RAG methods~\cite{graphrag,G-retriever,GRAG,wu2023retrieve,jin2024survey} demonstrated this potential by retrieving and integrating structured, relevant information from external sources, enabling LLMs to achieve superior performance on knowledge-intensive tasks.
However, 
these methods exhibit notable limitations when applied to complex problems.
Recent advancements have introduced methods like chain-of-thought prompting to enhance LLMs' reasoning on complex problems~\cite{Think_on_graph,tog2,graph_cot,chen2024plan,luo2025gfm,luo2024graph,li2024simple}.
Think-on-Graph~\cite{Think_on_graph} introduces a new approach that enables the LLM to iteratively execute beam search on the KGs, discover the most promising reasoning paths, and return the most likely reasoning results.
ToG-2~\cite{tog2} achieves deep and faithful reasoning in LLMs through an iterative knowledge retrieval process that involves collaboration between contexts and the KGs.
GRAPH-COT~\cite{graph_cot} propose a simple and effective framework to augment LLMs with graphs by encouraging LLMs to reason on the graph iteratively.
However, 
iterative reasoning processes are prone to error propagation, as errors cannot be corrected once introduced.
This leads to error accumulation and makes it difficult to reason holistically or revise previous steps. 
In contrast, our \ours\ framework constructs a complete reasoning plan in advance through top-down decomposition, forming a mind map that globally guides retrieval and reasoning.
None of the aforementioned approaches incorporate a self-verification phase. Once an incorrect inference is made, it propagates without correction. \ours\ addresses this by introducing a dual-LLM self-verification module inspired by dual-process cognitive theory, allowing the system to detect and revise incorrect outputs during the reasoning phase.
% , and each step compounds computational costs and processing time.

\section{Methodology}

In this section, we introduce \ours\, a human cognition-inspired graph-based retrieval-augmented generation framework designed to enhance complex reasoning in KGQA tasks. \ours\ simulates human cognitive strategies by decomposing complex questions into structured sub-problems, retrieving relevant knowledge in a hierarchical manner, and verifying reasoning results through dual-process reflection. The overall framework consists of three stages: (1) \emph{Decomposition}, (2) \emph{Structured Retrieval}, and (3) \emph{Reasoning with Self-Verification}. The workflow is illustrated in Figure~\ref{fg:overall}.

\subsection{Problem Formulation}
Given a natural language question $Q$, the goal is to generate a correct answer $A$ using an LLM $p_{\theta }$ augmented with a KG $\mathcal{G}$.
CogGRAG aim to design a graph-based RAG framework with an LLM backbone $p_{\theta }$ to enhance the complex problem-solving capabilities and generate the answer $A$.

\subsection{Top-Down Decomposition}
Inspired by human cognition in problem-solving, we simulate a top-down analytical process by decomposing the original question into a hierarchy of semantically coherent sub-questions. This decomposition process produces a \emph{reasoning mind map}, which serves as an explicit structure to guide subsequent retrieval and reasoning steps.
% In a mind map, each node denotes a sub-question; edges indicate reasoning dependencies.
% \ours\ decomposes $Q$ into a hierarchy of sub-questions forming a mind map $\mathcal{M}$. 
% Each node in $\mathcal{M}$ is a tuple $m = (q, t, s)$, where $q$ is a sub-question, $t$ is its level in the tree, and $s \in \{\texttt{Continue}, \texttt{End}\}$ indicates whether further decomposition is required.
In this mind map, each node represents a sub-question, and edges capture the logical dependencies among them. Specifically, \ours\ decomposes the original question $Q$ into a hierarchical structure forming a mind map $\mathcal{M}$. Each node in $\mathcal{M}$ is defined as a tuple $m = (q, t, s)$, where $q$ denotes the sub-question, $t$ indicates its depth level in the tree, and $s \in \{\texttt{Continue}, \texttt{End}\}$ specifies whether further decomposition is required.

The decomposition proceeds recursively using the LLM:
% \begin{equation} 
% \begin{split}
% & \left \{  \right  (q^{t+1}_{j}, s^{t+1}_{j}), j = 1,...,N \}  \\
% & = \text{Decompose}(q^{t}, p_{\theta }, prompt_{1}),
% \end{split}
% \end{equation}
\begin{equation}
\{(q^{t+1}_j, s^{t+1}_j)\}_{j=1}^{N} = \texttt{Decompose}(q^t, p_\theta, \texttt{prompt}_{dec}),
\end{equation}
where $N$ is determined adaptively by LLM. This process continues until all leaves in $\mathcal{M}$ are labeled with $\texttt{End}$, representing atomic questions.

This hierarchical planning mechanism allows CogGRAG to surface latent semantic dependencies that may be overlooked by traditional matching-based RAG methods. For instance, consider the query: ``\emph{The football manager who recruited David Beckham managed Manchester United during what timeframe?}'' A conventional retrieval pipeline may fail to resolve the intermediate entity ``Alex Ferguson,'' as it is not directly mentioned in the input question. In contrast, CogGRAG first decomposes the query into ``Who recruited David Beckham?'' and ``When did that manager coach Manchester United?'', thereby enabling the system to recover critical missing entities and construct a more accurate reasoning path.

By performing top-down question decomposition before any retrieval or reasoning, CogGRAG explicitly separates planning from execution. This design enables holistic reasoning over the entire problem space, mitigates error propagation, and sets the foundation for effective downstream processing.

\subsection{Structured Knowledge Retrieval}

Once the mind map $\mathcal{M}$ has been constructed through top-down decomposition, \ours\ enters retrieval stage. This phase is responsible for identifying the external knowledge required to support reasoning across all sub-questions. In contrast to prior iterative methods~\citep{graph_cot} that retrieve relevant content at each reasoning step, \ours\ performs a \textbf{one-pass, globally informed retrieval} guided by the full structure of $\mathcal{M}$. This design not only enhances retrieval completeness and contextual coherence, but also avoids error accumulation caused by sequential retrieval failures.

To retrieve relevant information from the KG $\mathcal{G}$, \ours\ extracts two levels of key information from $\mathcal{M}$.
\textbf{Local-level information} captures entities and fine-grained relational facts associated with individual sub-questions. These include entity mentions, entity-relation pairs, and triples.
\textbf{Global-level information} reflects semantic dependencies across multiple sub-questions and is expressed as interconnected subgraphs. These subgraphs encode logical chains or joint constraints necessary to resolve the overall problem.
For example, given the decomposed question ``\emph{Which football manager recruited David Beckham?}'', we can extract local-level information such as the entity (David Beckham) and the triple (manager, recruited, David Beckham).
% By further analyzing another decomposed question ``\emph{During which years did that manager lead Manchester United?}'', we can derive global-level information, represented as the subgraph [(manager, recruited, David Beckham), (manager, manage, Manchester United)].
By additionally considering another decomposed question—such as ``\emph{During which years did that manager lead Manchester United?}''—we can derive global level information, which is represented in the form of a subgraph [(manager, recruited, David Beckham), (manager, manage, Manchester United)].
The extraction process is performed using the LLM:
\begin{equation}
    \mathcal{K} = \texttt{Extract}(\mathcal{M}, p_\theta, \texttt{prompt}_{ext}),
\end{equation}
where $\mathcal{K}$ denotes the set of all extracted keys, including entity nodes, triples, and subgraph segments. This design ensures that both isolated facts and relational contexts are captured prior to KGs querying.
These keys guide graph-based retrieval. For each entity $e$ in the extracted key set $\mathcal{K}$, we expand to its neighborhood in $\mathcal{G}$, yielding a candidate triple set $\tilde{\mathcal{T}}$. A semantic similarity filter is applied to retain only relevant triples:
\begin{equation}
    \mathcal{T} = \{ \tau \in \tilde{\mathcal{T}}, k \in \mathcal{K} \mid \texttt{sim}(\tau, k) > \varepsilon \},
\end{equation}
where $\texttt{sim}(\cdot)$ is cosine similarity, $\varepsilon$ is a threshold,
$\tau $ is the triple in candidate triple
set $\tilde{\mathcal{T}}$
and $k$ is the triple in the extracted key set $\mathcal{K}$.
The final retrieved triple set $\mathcal{T}$ serves as the input evidence pool for the reasoning module. By leveraging both local and global context from $\mathcal{M}$, \ours\ ensures that $\mathcal{T}$ is semantically rich, structurally connected, and targeted toward the entire reasoning path.

\subsection{Reasoning with Self-Verification}
The final stage of \ours\ emulates human problem-solving behavior, where conclusions are not only derived through reasoning but also reviewed through self-reflection~\cite{frederick2005cognitive}. To this end, \ours\ employs a \textbf{dual-process reasoning} architecture, composed of two distinct LLMs:
\begin{itemize}
    \item \textbf{LLM\textsubscript{res}}: responsible for answer generation via bottom-up reasoning over the mind map $\mathcal{M}$;
    \item \textbf{LLM\textsubscript{ver}}: responsible for evaluating the validity of generated answers based on context and prior reasoning history.
\end{itemize}
This dual-agent setup is inspired by dual-process theories in cognitive psychology~\cite{vaisey2009motivation}, where System 1 performs intuitive reasoning and System 2 monitors and corrects errors.

\paragraph{Bottom-Up Reasoning.}
Given the final retrieved triple set $\mathcal{T}$ and the structured mind map $\mathcal{M}$, \ours\ performs reasoning in a bottom-up fashion: sub-questions at the lowest level are answered first, and their verified results are recursively used to resolve higher-level nodes.
% Let $\hat{\mathcal{M}} $ denote the set of sub-nodes that have been (1) answered through bottom-up reasoning and (2) verified as correct by the self-verification module. Each element in $\hat{\mathcal{M}}$ is a tuple $(q, \hat{a})$, where $q$ is a sub-question and $\hat{a}$ is its verified answer. 
% The verified subgraph $\hat{\mathcal{M}}$ serves as trusted intermediate results to support the reasoning of higher-level sub-questions in the hierarchical structure.

Let $\mathcal{M}$ denote the full mind map. We define $\hat{\mathcal{M}} \subset \mathcal{M}$ as the subset of sub-questions that have been answered and verified. Each element in $\hat{\mathcal{M}}$ is a tuple $(q, \hat{a})$, where $q$ is a sub-question and $\hat{a}$ is its verified answer.
% Let $q^t$ denote a sub-question at level $t$.
% , and $\hat{\mathcal{M}}$ denote the set of verified answers.
Then the reasoning LLM generates a candidate answer:
\begin{equation}
    a^t = \texttt{LLM}_{\text{res}}(\mathcal{T}, q^t, \hat{\mathcal{M}}, \texttt{prompt}_{res}).
\end{equation}

\paragraph{Self-Verification and Re-thinking.}
The candidate answer $a^t$ is passed to the verifier LLM along with the current reasoning path. The verifier assesses consistency, factual grounding, and logical coherence. If the answer fails validation, a re-thinked response $\hat{a}^t$ is regenerated:
% \begin{equation}
%     \hat{a}_t =
%     \begin{cases}
%     \texttt{LLM}_{\text{res}}(\mathcal{T}, q^t, \hat{\mathcal{M}}_{>t}, \texttt{prompt}_3), & \text{if } \texttt{LLM}_{\text{ver}}(a_t) = \texttt{False}, \\
%     a_t, & \text{otherwise}.
%     \end{cases}
% \end{equation}
\begin{equation}
\hat{a}^t = 
\begin{cases}
\texttt{LLM}_{\text{res}}(\mathcal{T}, q^t, \hat{\mathcal{M}}, \texttt{prompt}_{rethink}) & \text{if } \delta^t = \texttt{False}, \\
a^t & \text{otherwise.}
\end{cases}
\end{equation}
Here, $\delta^t$ is a boolean indicator defined as:
\begin{equation}
\delta^t = \texttt{LLM}_{\text{ver}}(q^t, a^t, \hat{\mathcal{M}}, \texttt{prompt}_{ver}),
\end{equation}
which returns \texttt{True} if the verification module accepts $a^t$, and \texttt{False} otherwise.
In addition to verification, \ours\ incorporates a selective abstention mechanism. When the reasoning LLM cannot confidently produce a grounded answer based on $\mathcal{T}$, it is explicitly prompted to respond with ``\texttt{I don't know}'' rather than hallucinating.
% The final answer $A$ to the original question $Q$ is obtained by recursively aggregating verified answers $\hat{a}^t$ from leaf nodes up to the root node $q^0$:
% \begin{equation}
% A = \hat{a}^0.
% \end{equation}
The final answer $A$ to the original question $Q$ is obtained by recursively aggregating verified answers $\hat{a}^t$ from the leaf nodes up to the root node $q^0$, such that $A = \hat{a}^0$.

This bottom-up reasoning pipeline, governed by a globally constructed mind map and protected by verification layers, enables \ours\ to perform accurate and reliable complex reasoning over structured knowledge.
All prompt templates used in \ours\ can be found in the Appendix.

\begin{table*}[h]

% \vspace{-1em}
\centering
% \fontsize{5.0pt}{5.0pt}\selectfont
% \scriptsize
\small
% \resizebox{1.0\textwidth}{!}{
\begin{tabular}{lp{1.8cm}p{1cm}p{1cm}p{1cm}p{1cm}p{1cm}p{1cm}p{1cm}p{1cm}p{1cm}}
\toprule
\multirow{2}{*}{Type} & \multirow{2}{*}{Method} 
& \multicolumn{3}{c}{HotpotQA} 
& \multicolumn{3}{c}{CWQ}
& \multicolumn{3}{c}{WebQSP} \\
\cmidrule(l){3-11} 
& & RL & EM & F1 & RL & EM & F1 & RL & EM & F1 \\
\midrule
\multirow{2}{*}{LLM-only} 
& Direct       & 19.1\% & 17.3\% & 18.7\% & 31.4\% & 28.8\% & 31.7\% & 51.4\% & 47.9\% & 53.5\% \\
& CoT          & 23.3\% & 20.8\% & 22.1\% & 35.1\% & 32.7\% & 33.5\% & 55.2\% & 51.6\% & 55.3\% \\
\midrule
\multirow{2}{*}{LLM+KG}    
& Direct+KG    & 27.5\% & 23.7\% & 27.6\% & 39.7\% & 35.1\% & 38.3\% & 52.5\% & 49.3\% & 52.1\% \\
& CoT+KG       & 28.7\% & 25.4\% & 26.9\% & 42.2\% & 37.6\% & 40.8\% & 52.8\% & 48.1\% & 50.5\% \\
\midrule
\multirow{4}{*}{Graph-based RAG} 
& ToG          & 29.3\% & 26.4\% & 29.6\% & 49.1\% & 46.1\% & 47.7\% & 54.6\% & 57.4\% & 56.1\% \\
& MindMap      & 27.9\% & 25.6\% & 27.8\% & 46.7\% & 43.7\% & 44.1\% & 56.6\% & 53.1\% & 58.3\% \\
& RoG          & 30.7\% & 28.1\% & 30.4\% & 55.3\% & 51.8\% & 54.7\% & 65.2\% & \textbf{62.8\%} & \textbf{67.2\%} \\
& GoG          & 31.5\% & 30.1\% & 31.1\% & 55.7\% & 52.4\% & 54.8\% & \textbf{65.5\%} & 59.1\% & 63.6\% \\
\midrule
\textbf{Ours} & \textbf{CogGRAG} 
              & \textbf{34.4\%} & \textbf{30.7\%} & \textbf{35.5\%} 
              & \textbf{56.3\%} & \textbf{53.4\%} & \textbf{55.8\%} 
              & 59.8\% & 56.1\% & 58.9\% \\
\bottomrule
\end{tabular}
% }
\caption{Overall results of our \ours\ on three KBQA datasets.
The best score on each dataset is highlighted.
}
\label{tb:main1}
\end{table*}

\section{Experiments}
\subsection{Experimental Settings}

\noindent \textbf{Datasets and evaluation metrics.}
In order to test \ours's complex problem-solving capabilities on KGQA tasks, we evaluate \ours\ on three widely used complex KGQA benchmarks: 
(1) \textbf{HotpotQA}~\cite{yang2018hotpotqa},
(2) \textbf{WebQSP}~\cite{yih2016value},
(3) \textbf{CWQ}~\cite{talmor2018web}.
Following previous work~\cite{cok}, full Wikidata~\cite{wikidata} is used as structured knowledge sources for all of these datasets.
Considering that Wikidata is commonly used for training LLMs, there is a need for a domain-specific QA dataset that is not exposed during the pretraining process of LLMs in order to better evaluate the performance.
Thus, we also test \ours\ on a recently released domain-specific dataset \textbf{GRBENCH}~\cite{graph_cot}.
All methods need to interact with domain-specific graphs containing rich knowledge to solve the problem in this dataset.
The statistics and details of these datasets can
be found in Appendix.
For all datasets,
we use three evaluation metrics:
(1) \textbf{Exact match (EM):} measures whether the predicted answer or result matches the target answer exactly.
(2) \textbf{Rouge-L (RL):} measures the longest common subsequence of words between the responses and the ground truth answers.
(3) \textbf{F1 Score (F1):} computes the harmonic mean of precision and recall between the predicted and gold answers, capturing both completeness and exactness of overlapping tokens.

\noindent\textbf{Baselines.} 
In our main results, we compare \ours\ with three types state-of-the-art methods:
(1) \textbf{LLM-only} methods without external knowledge, including direct reasoning and CoT~\cite{prompt_cot} by LLM.
(2) \textbf{LLM+KG} methods integrate relevant knowledge retrieved from the KG into the LLM to assist in reasoning, 
including direct reasoning and CoT by LLM.
(3) \textbf{Graph-based RAG methods} allow KGs and LLMs to work in tandem, complementing each other’s capabilities at each step of graph reasoning, including Mindmap~\cite{mindmap}, Think-on-graph~\cite{Think_on_graph}, Graph-CoT~\cite{graph_cot}, 
RoG~\cite{luo2024rog},
GoG~\cite{xu2024gog}.
% and  G-retriever~\cite{he2024gretriever}.
Due to the space limitation, we move details on datasets, baselines and experimental setup to Appendix.

\noindent\textbf{Experimental Setup.}
% \label{sec:ex_setup}
We conduct experiments with four LLM backbones: LLaMA2-13B~\cite{llama2}, LLaMA3-8B~\cite{llama3}, Qwen2.5-7B~\cite{qwen2} and Qwen2.5-32B~\cite{hui2024qwen2}.
For all LLMs, we load the checkpoints from huggingface\footnote{\url{https://huggingface.co}} and use the models directly without fine-tuning.
We implemented \ours\ and conducted the experiments with one A800 GPU.
Consistent with the Think-on-graph settings, we set the temperature parameter to 0.4 during exploration and 0 during reasoning.
The threshold $\epsilon$ in the retrieval step is set to $0.7$.

\begin{table*}[]
\centering
\small
% \resizebox{1.0\textwidth}{!}{
\begin{tabular}{p{2.1cm} p{3.7cm}ccccccccc}
\toprule
\multirow{2}{*}{Type} & \multirow{2}{*}{Method} 
& \multicolumn{3}{c}{HotpotQA}      
& \multicolumn{3}{c}{CWQ}           
& \multicolumn{3}{c}{WebQSP}        \\ 
\cmidrule(l){3-11} 
 & & RL & EM & F1 & RL & EM & F1 & RL & EM & F1 \\
\midrule

\multirow{4}{*}{LLM-only}        
& Qwen2.5-7B        & 15.3\% & 15.0\% & 15.8\% & 25.4\% & 24.1\% & 24.5\% & 46.7\% & 45.3\% & 45.5\% \\
& LLaMA3-8B         & 17.5\% & 14.9\% & 16.2\% & 30.3\% & 27.5\% & 29.0\% & 50.4\% & 45.1\% & 48.3\% \\
& LLaMA2-13B        & 19.1\% & 17.3\% & 18.7\% & 31.4\% & 28.8\% & 31.7\% & 51.4\% & 47.9\% & 53.5\% \\ 
& Qwen2.5-32B       & 28.7\% & 27.4\% & 28.5\% & 55.1\% & 50.3\% & 54.2\% & 68.4\% & 60.5\% & 65.1\% \\ 

\midrule

\multirow{4}{*}{LLM+KG}          
& Qwen2.5-7B+KG     & 24.2\% & 15.6\% & 21.4\% & 33.8\% & 32.1\% & 34.9\% & 46.7\% & 45.3\% & 46.1\% \\
& LLaMA3-8B+KG      & 25.9\% & 21.4\% & 23.6\% & 40.6\% & 35.3\% & 39.1\% & 53.6\% & 49.1\% & 52.3\% \\
& LLaMA2-13B+KG     & 27.5\% & 23.7\% & 27.6\% & 39.7\% & 35.1\% & 38.3\% & 52.5\% & 49.3\% & 52.1\% \\ 
& Qwen2.5-32B+KG    & 35.6\% & 32.8\% & 34.9\% & 58.3\% & 54.7\% & 57.6\% & 70.2\% & 65.1\% & 68.4\% \\ 

\midrule

\multirow{4}{*}{Graph-Based RAG} 
& \ours\ w/ Qwen2.5-7B   & 28.4\% & 27.1\% & 28.2\% & 50.5\% & 45.7\% & 48.9\% & 53.2\% & 51.6\% & 55.0\% \\
& \ours\ w/ LLaMA3-8B    & 32.1\% & 27.2\% & 31.0\% & 53.5\% & 48.4\% & 52.6\% & 57.2\% & 55.3\% & 55.4\% \\
& \ours\ w/ LLaMA2-13B   & 34.4\% & 30.7\% & 35.5\% & 56.3\% & 53.4\% & 55.8\% & 59.8\% & 56.1\% & 58.9\% \\ 
& \ours\ w/ Qwen2.5-32B  & 40.5\% & 37.1\% & 40.2\% & 66.5\% & 62.7\% & 65.4\% & 74.1\% & 68.3\% & 73.0\% \\ 

\bottomrule
\end{tabular}
% }
\caption{Overall results of our \ours\ with different backbone models on three KBQA datasets.
}
\label{ta:backbone}
\end{table*}

\subsection{Main Results}

We perform experiments to verify the effectiveness of our framework \ours, and report the results in Table~\ref{tb:main1}.
We use Rouge-L (RL), Exact match (EM) and F1 Score (F1) as metric for all three datasets.
The backbone model for all the methods is LLaMA2-13B.
From the table, the following observations can be derived:
(1) \ours\ achieves the best results in most cases except on WebSQP.
Since the dataset is widely used, 
we attribute the reason to be data leakage.
(2) Compared to methods that incorporate external knowledge, the LLM-only approach demonstrates significantly inferior performance. 
This performance gap arises
% because LLMs lack the necessary knowledge for reasoning tasks,
from the lack of necessary knowledge in LLMs for reasoning tasks,
highlighting the critical role of external knowledge integration in enhancing the reasoning capabilities of LLMs.
(3) Graph-based RAG methods demonstrate superior performance compared to LLM+KG approaches. 
This performance advantage is particularly evident in complex problems, where not only external knowledge integration but also
involving ``thinking procedure''
is essential. 
These methods synergize LLMs with KGs to enhance performance by retrieving and reasoning over the KGs,
thereby generating the most probable inference outcomes through
``thinking''
on the KGs.
% (4) \ours\ demonstrates the best results in most cases, 

% \vspace{-0.1cm}
We attribute \ours~'s outstanding effectiveness in most cases
primarily to its ability to decompose complex problems and construct a structured mind map prior to retrieval. 
This approach enables the 
construction
% establishment
of a comprehensive reasoning pathway, facilitating more precise and targeted retrieval of relevant information. 
Furthermore, \ours\ incorporates a self-verification mechanism during the reasoning phase, further enhancing the accuracy and reliability of the final results. 
% All the results effectiveness of our proposed method in addressing complex problems.
Together, these designs collectively enhance LLMs' ability to tackle complex problems.

\subsection{Performance with different backbone models}
We evaluate how different backbone models affect its
performance on three datasets HotpotQA, CWQ and WebQSP, 
and report the results in Table~\ref{ta:backbone}.
We conduct experiments with three LLM backbones LLaMA2-13B, LLaMA3-8B, Qwen2.5-7B and Qwen2.5-32B.
For all LLMs, we load the checkpoints from huggingface and use the models directly without fine-tuning.
From the table, we can observe the following key findings:
(1) \ours\ achieves the best results across all backbone models compared to the baseline approaches, demonstrating the robustness and stability of our method.
(2) The performance of our method improves consistently as the model scale increases, reflecting enhanced reasoning capabilities. This trend suggests that our approach has significant potential for further exploration with larger-scale models, indicating promising scalability and adaptability.

\begin{table*}[]
\centering
\small
% \fontsize{6pt}{6.5pt}\selectfont
% \resizebox{1.0\textwidth}{!}{
\begin{tabular}{lcccccccccccc}
\toprule
\multirow{2}{*}{Method}            
& \multicolumn{3}{c}{E-commerce} 
& \multicolumn{3}{c}{Literature} 
& \multicolumn{3}{c}{Academic} 
& \multicolumn{3}{c}{Healthcare} 
\\ \cmidrule(l){2-13} 
& RL & EM & F1 
& RL & EM & F1 
& RL & EM & F1 
& RL & EM & F1 
\\ \midrule

LLaMA2-13B                           
& 7.1\%  & 6.8\%  & 6.9\%  
& 5.4\%  & 5.1\%  & 5.3\%  
& 5.4\%  & 4.7\%  & 5.1\%  
& 4.3\%  & 3.1\%  & 3.6\%  
\\

Graph-CoT                            
& 26.4\% & 24.0\% & 25.3\%  
& 26.7\% & 23.3\% & 24.9\%  
& 19.3\% & 14.8\% & 16.9\%  
& \textbf{28.1\%} & 25.2\% & \textbf{26.7\%}  
\\

\ours                               
& \textbf{30.2\%} & \textbf{28.7\%} & \textbf{29.5\%}  
& \textbf{32.4\%} & \textbf{30.1\%} & \textbf{31.3\%}  
& \textbf{23.6\%} & \textbf{21.5\%} & \textbf{22.7\%}  
& 27.4\% & \textbf{25.6\%} & 26.2\%  
\\ 

\bottomrule
\end{tabular}
% }
\caption{Overall results of our \ours\ on GRBENCH dataset.
We highlight the best score on each dataset in bold.
}
\label{tb:grbench}
\end{table*}

\subsection{Performance on domain-specific KG}
% Given the potential risks associated with using Wikidata in LLMs training,
Given the risk of data leakage due to Wikidata being used as pretraining corpora for LLMs,
we further evaluate the performance of \ours\ on a
recently released
domain-specific dataset GRBENCH~\cite{graph_cot} and report the results in Table~\ref{tb:grbench}. 
This dataset requires all questions to interact with a domain-specific KG.
We use Rouge-L (RL), Exact match (EM) and F1 Score (F1) as metrics for this dataset.
The backbone model for all the methods is LLaMA2-13B~\cite{llama2}.
The table reveals the following observations:
(1) \ours\ continues to outperform in most cases. 
This demonstrates that our method consistently achieves stable and reliable results even on domain-specific KGs.
(2) Both \ours\ and Graph-CoT outperform LLaMA2-13B by more than 20\%, 
% a significant difference attributed to 
which can be ascribed to
the fact that LLMs are typically not trained on domain-specific data. 
In contrast, graph-based RAG methods can effectively supplement LLMs with external knowledge, thereby enhancing their reasoning capabilities. 
This result underscores the effectiveness of the RAG approach in bridging knowledge gaps and improving performance in specialized domains.

\subsection{Ablation Study}
\begin{figure}[]
  \centering  \includegraphics[width=0.91\linewidth]{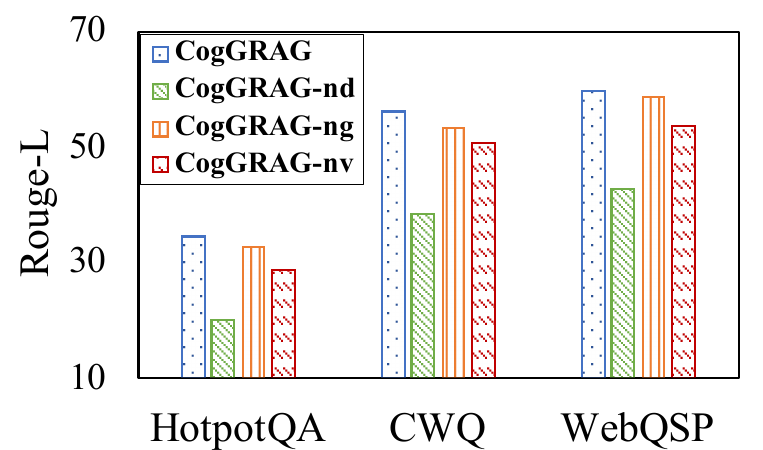}
  \caption{Ablation study on the main components of \ours.}
  \label{fg:ablation}
\end{figure}

The ablation study is conducted to understand the importance of main components of \ours.
We select HotpotQA, CWQ and WebQSP as three representative datasets.
% To evaluate the contribution of each component in our framework, we conducted a series of ablation studies. 
First, we remove the problem decomposition module, directly extracting information for the target question, and referred to this variant as \ours-nd (\textbf{n}o \textbf{d}ecomposition). Next, we eliminate the global-level retrieval phase, naming this variant \ours-ng (\textbf{n}o \textbf{g}lobal level). 
Finally, we remove the self-verification mechanism during the reasoning stage, designating this variant as \ours-nv (\textbf{n}o \textbf{v}erification).
These experiments aim to systematically assess the impact of each component on the overall performance of the framework.
We compare \ours\ with these three variants, and the results are presented in Figure~\ref{fg:ablation}.
Our findings show that \ours\ outperforms all the variants on the three datasets.
Furthermore, the performance gap between \ours\ and \ours-nd highlights the importance of decomposition for complex problem reasoning in KGQA tasks.

\begin{table}[]
\centering
\small
% \resizebox{0.48\textwidth}{!}{
\begin{tabular}{@{}lccc@{}}
\toprule
Method               & Correct ($\uparrow$) & Missing ($\uparrow$) & Hallucination ($\downarrow$) \\ \midrule
LLaMA2-13B           & 19.1\%  & 25.7\%  & 55.2\%        \\
ToG                  & 29.3\%  & 20.2\%  & 50.5\%        \\
MindMap              & 27.9\%  & 22.4\%  & 49.7\%        \\
\ours\ & \textbf{34.4\%}  & \textbf{40.6\%}  & \textbf{25.0\%}        \\ \bottomrule
\end{tabular}
% }
\caption{Overall results of our \ours\ on GRBENCH dataset.
We highlight the best score on each dataset in bold.
}
\label{tb:missing}
\end{table}

\subsection{Hallucination and Missing Evaluation}
In the reasoning and self-verification phase of \ours, we prompt the LLM to respond with ``I don't know'' when encountering questions with insufficient or incomplete relevant information during the reasoning process. 
This approach is designed to mitigate the hallucination issue commonly observed in LLMs. 
To evaluate the effectiveness of this strategy, we test the model on the HotpotQA dataset, with results reported in Table~\ref{tb:missing}. 
We categorize the responses into three types: ``Correct'' for accurate answers, ``Missing'' for cases where the model responds with ``I don't know,'' and ``Hallucination'' for incorrect answers. 
As shown in the table results, our model demonstrates the ability to refrain from answering questions with insufficient information, significantly reducing the occurrence of hallucinations. 
% This capability highlights the effectiveness of our approach in enhancing the
% % reliability and accuracy 
% \zhu{reliability and truthfulness} of the model's responses.

\begin{table}[]
\centering
\small
% \resizebox{0.45\textwidth}{!}{
\begin{tabular}{lcccc}
\toprule
\textbf{Method} & \textbf{Academic} & \textbf{Amazon} & \textbf{Goodreads} & \textbf{Disease} \\
\midrule
Graph-CoT     & 22.39 & 29.04 & 15.79 & 32.97 \\
GoG           & 21.50 & 32.20 & 18.80 & 37.50 \\
% G-Retriever   & 16.70 & 26.90 & 14.50 & 30.10 \\
CogGRAG       & 18.24 & 31.57 & 16.32 & 35.64 \\
\bottomrule
\end{tabular}
% }
\caption{Average inference time per question (s).
}
\label{tb:times}
\end{table}

\subsection{Performance in Inference Times}
% CogGRAG prompts the LLM for decomposition, key information extraction, reasoning, and self-verification. 
We conduct the verification on four datasets GRBENCH-Academic, GRBENCH-Amazon, GRBENCH-Goodreads, and GRBENCH-Disease. Table~\ref{tb:times} presents the average runtime of CogGRAG, Graph-CoT and GoG.
From the table, we can see that CogGRAG is on par with Graph-CoT and GoG.
% performs faster than Graph-CoT, while it is on par with other baselines on the other three datasets. 
Although our reasoning process introduces self-verification, which increases the time cost, CogGRAG does not require iterative repeated reasoning and retrieval. 
Instead, it retrieves all relevant information in one step based on the mind map, avoiding redundant retrieval, which provides an advantage in large-scale KGs. Moreover, while self-verification adds additional time costs, it also improves the accuracy of reasoning results, as confirmed in the ablation experiments.

\section{Conclusion}

In this paper, we proposed CogGRAG, a graph-based RAG framework to enhance LLMs’ complex reasoning for KGQA tasks. CogGRAG generates tree-structured mind maps, explicitly encoding semantic relationships among subproblems, which guide both local and global knowledge retrieval. A self-verification module further detects and revises potential errors, forming a dual-phase reasoning paradigm that reduces hallucinations and improves factual consistency. Extensive experiments show that CogGRAG substantially outperforms existing RAG baselines on complex multi-hop QA benchmarks.

\section{Acknowledgments}
This work was supported in part by the National Natural Science Foundation of China under Grant 42375146 and National Natural Science Foundation of China No. 62202172.

\bibliography{aaai2026}

\newpage
\appendix

\section{Datastes}
\label{sec:data}
Here, we introduce the four datasets used in our experiments in detail.
For HotpotQA and CWQ datasets, we evaluated the performance of all methods on the dev set.
For WebQSP dataset, we evaluated the performance of all methods on the train set.
The statistics and details of these datasets can be found in Table~\ref{statistics_dataset}, and we describe each dataset in detail below:

\noindent\textbf{\emph{HotpotQA}} is a large-scale question answering dataset aimed at facilitating the development of QA systems capable of performing explainable, multi-hop reasoning over diverse natural language. It contains 113k question-answer pairs that were collected through crowdsourcing based on Wikipedia articles.

\noindent\textbf{\emph{WebQSP}} is a dataset designed specifically for question answering and information retrieval tasks, aiming to promote research on multi-hop reasoning and web-based question answering techniques.
It contains 4,737 natural language questions that are answerable using a subset Freebase KG~\cite{bollacker2008freebase}.

\noindent\textbf{\emph{CWQ}} is a dataset specially designed to evaluate the performance of models in complex question answering tasks. It is generated from WebQSP by extending the question entities or adding constraints to answers, in order to construct more complex multi-hop questions. 

\noindent\textbf{\emph{GRBENCH}} is a dataset to evaluate how effectively LLMs can interact with domain-specific graphs containing rich knowledge to solve the desired problem. GRBENCH contains 10 graphs from 5 general domains (academia, e-commerce, literature, healthcare, and legal). Each data sample in GRBENCH is a question-answer pair. 

\begin{table}[h]
\centering
% \tiny
\resizebox{0.48\textwidth}{!}{
\begin{tabular}{@{}ccccc@{}}
\toprule
\multirow{2}{*}{Domain}    & \multirow{2}{*}{Dataset} & \multicolumn{3}{c}{Data Split} \\ \cmidrule(l){3-5} 
                           &                          & Train      & Dev     & Test    \\ \midrule
\multirow{3}{*}{Wikipedia} & HotpotQA                 & 90564      & 7405    & 7405    \\
                           & WebQSP                   & 3098       & 0       & 1639    \\
                           & CWQ                      & 27,734     & 3480    & 3475    \\ \midrule
Academic                   & GRBENCH-Academic         & \multicolumn{3}{c}{850}        \\
E-commerce                 & GRBENCH-Amazon           & \multicolumn{3}{c}{200}        \\
Literature                 & GRBENCH-Goodreads        & \multicolumn{3}{c}{240}        \\
Healthcare                 & GRBENCH-Disease          & \multicolumn{3}{c}{270}        \\ \bottomrule
\end{tabular}
}
\caption{Datasets statistics.
}
\label{statistics_dataset}
\end{table}

\begin{figure*}[h]
  \centering  \includegraphics[width=0.9\linewidth]{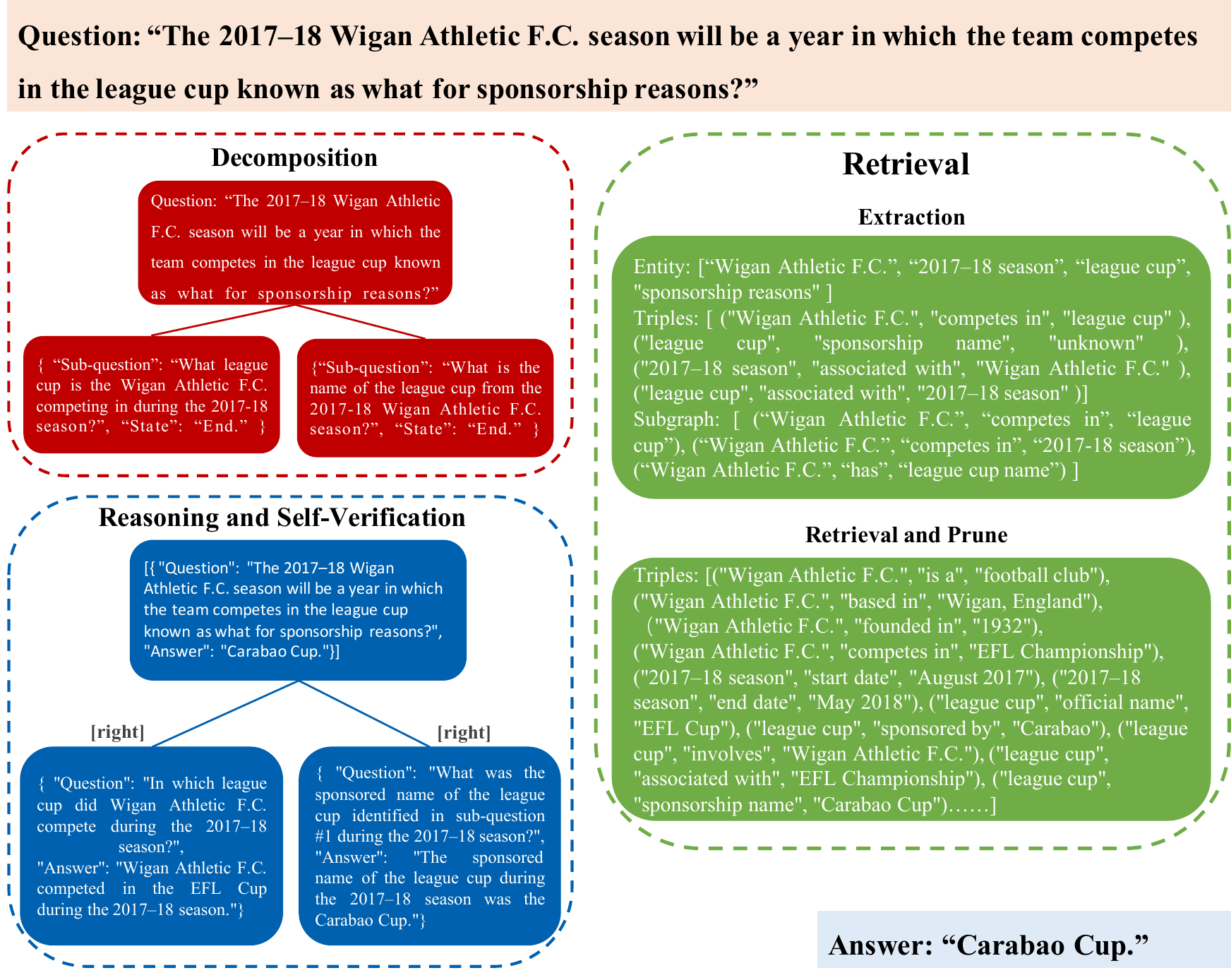}
  \caption{Case of \ours.}
  \label{fg:case1}
\end{figure*}

\section{Baselines}
\label{sec:baseline}
In this subsection, we introduce the baselines used in our experiments, including LLaMA2-13B, LLaMA3-8B, Qwen2.5-7B, Qwen2.5-32B, Chain-of-Thought (CoT) prompting, Think-on-graph (ToG), MindMap, Graph-CoT, Reasoning on Graphs (RoG), Generate-on-Graph (GoG).
% and G-retriever.

\noindent\textbf{\emph{LLaMA2-13B}}~\cite{llama2} is a member of the LLM series developed by Meta and is the second generation version of the LLaMA (Large Language Model Meta AI) model. LLaMA2-13B contains 13 billion parameters and is a medium-sized large language model.

\noindent\textbf{\emph{LLaMA3-8B}}~\cite{llama3} is an efficient and lightweight LLM with 8 billion parameters, designed to provide high-performance natural language processing capabilities while reducing computing resource requirements.

\noindent\textbf{\emph{Qwen2.5-7B}}~\cite{qwen2} is an efficient and lightweight LLM launched by Alibaba with 7 billion parameters. It aims at making it more efficient, specialized, and optimized for particular applications.

\noindent\textbf{\emph{Qwen2.5-32B}}~\cite{hui2024qwen2} is an efficient LLM launched by Alibaba with 32 billion parameters. It is a more comprehensive foundation for real-world applications such as Code Agents. Not only enhancing coding capabilities but also maintaining its strengths in mathematics and general competencies.

\noindent\textbf{\emph{Chain-of-Thought}}~\cite{prompt_cot} is a method designed to enhance a model's reasoning ability, particularly for complex tasks that require multiple reasoning steps. It works by prompting the model to generate intermediate reasoning steps rather than just providing the final answer, thereby improving the model’s ability to reason through complex problems.

\noindent\textbf{\emph{Think-on-graph}}~\cite{Think_on_graph} proposed a new LLM-KG integration paradigm, ``LLM $\otimes $ KG'', which treats the LLM as an agent to interactively explore relevant entities and relationships on the KG and perform reasoning based on the retrieved knowledge. It further implements this paradigm by introducing a novel approach called ``Graph-based Thinking'', in which the LLM agent iteratively performs beam search on the KG, discovers the most promising reasoning paths, and returns the most likely reasoning result.

\noindent\textbf{\emph{MindMap}}~\cite{mindmap} propose a novel prompting pipeline, that leverages KGs to enhance LLMs’ inference and transparency. It enables LLMs to comprehend KG inputs and infer with a combination of implicit and external knowledge. Moreover, it elicits the mind map of LLMs, which reveals their reasoning pathways based on the ontology of knowledge. 

\noindent\textbf{\emph{Graph-CoT}}~\cite{graph_cot} propose a simple and effective framework called Graph Chain-of-thought to augment LLMs with graphs by encouraging LLMs to reason on the graph iteratively. 
Moreover, it manually construct a Graph Reasoning Benchmark dataset called GRBENCH, containing 1,740 questions that can be answered with the knowledge from 10 domain graphs.

\noindent\textbf{\emph{Reasoning on Graphs}}~\cite{luo2024rog} propose a novel
method called reasoning on graphs (RoG) that synergizes LLMs with KGs to enable faithful and interpretable reasoning. Specifically, they present a planning retrieval-reasoning framework, where RoG first generates relation paths grounded
by KGs as faithful plans. These plans are then used to retrieve valid reasoning paths from the KGs for LLMs to conduct faithful reasoning.

\noindent\textbf{\emph{Generate-on-Graph}}~\cite{xu2024gog} propose leveraging LLMs for QA under Incomplete Knowledge Graph (IKGQA), where the provided KG lacks some of the factual triples for each question, and construct corresponding datasets. To handle IKGQA, they propose a
training-free method called Generate-on-Graph
(GoG), which can generate new factual triples
while exploring KGs.

\section{Case Studies}
\label{sec:case}
In this section, we present a case analysis of the HotpotQA dataset to evaluate the performance of \ours. As illustrated in Figure~\ref{fg:case1}, \ours\ decomposes the input question into two logically related sub-questions. 
Specifically, the complex question is broken down into ``In which league cup'' and ``What was the sponsored name'', allowing the system to first identify the league cup and then determine its sponsored name based on the identified league.
These sub-questions form a mind map, capturing the relationships between different levels of the problem. 
Next, \ours\ extracts key information from all sub-questions, including both local level information within individual sub-questions and global level information across different sub-questions. 
A subgraph is constructed to represent the interconnected triples within the subgraph, enabling a global perspective to model the relationships between different questions. 
The retrieved triples are pruned based on similarity metrics. 
All information retrieved from the KG is represented in the form of triples.
Using this knowledge, \ours\ prompts the LLM to perform bottom-up reasoning and self-verification based on the constructed mind map.
Through this process, the system ultimately derives the answer to the target question: ``Carabao Cup.'' This case demonstrates the effectiveness of \ours\ in handling complex, multi-step reasoning tasks by leveraging hierarchical decomposition, structured knowledge retrieval, and self-verification mechanisms.

\begin{figure*}[h]
  \centering  \includegraphics[width=0.9\linewidth]{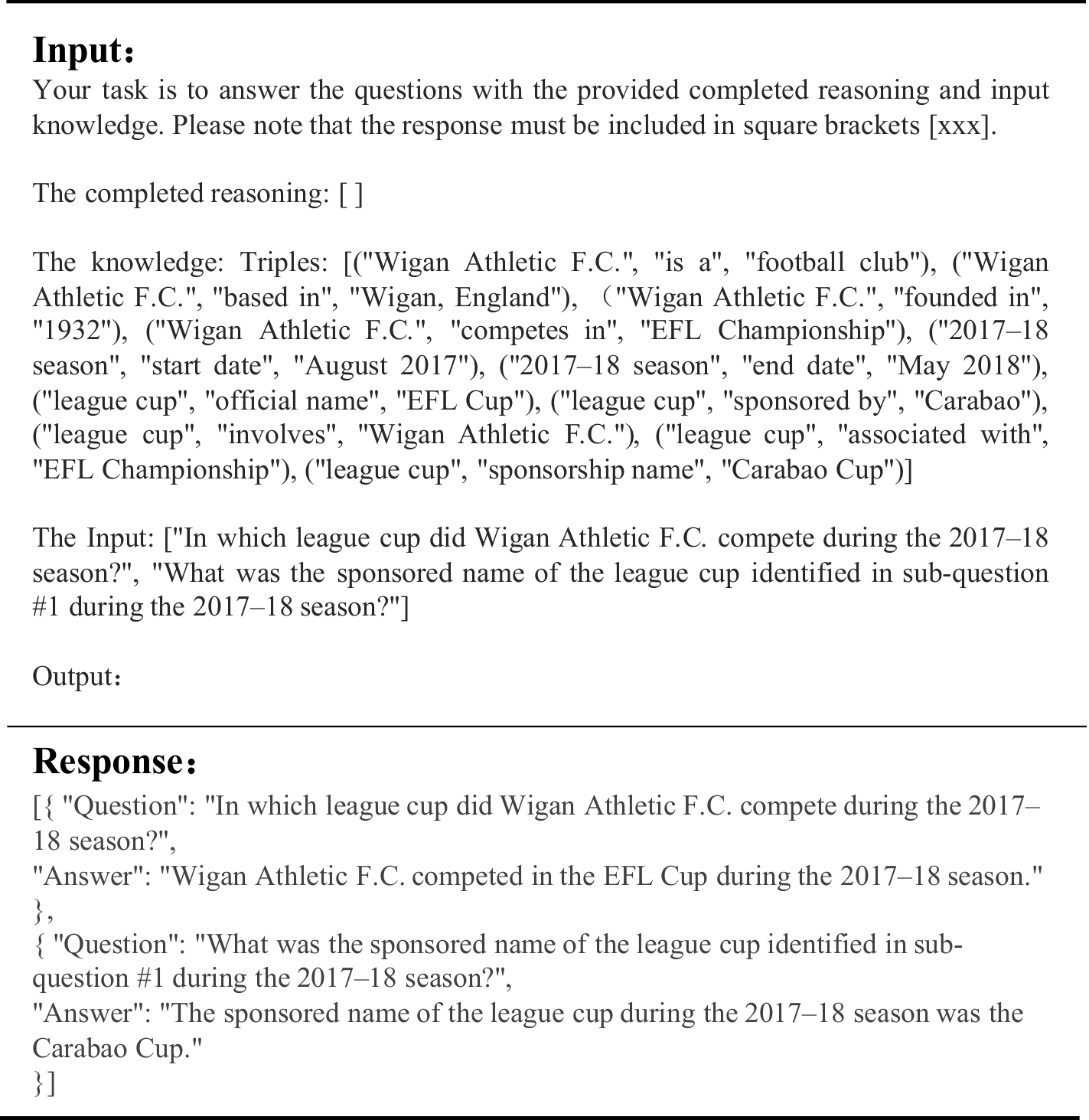}
  \caption{The prompt case of reasoning.}
  \label{fg:pcase1}
\end{figure*}

\begin{figure*}[h]
  \centering  \includegraphics[width=0.9\linewidth]{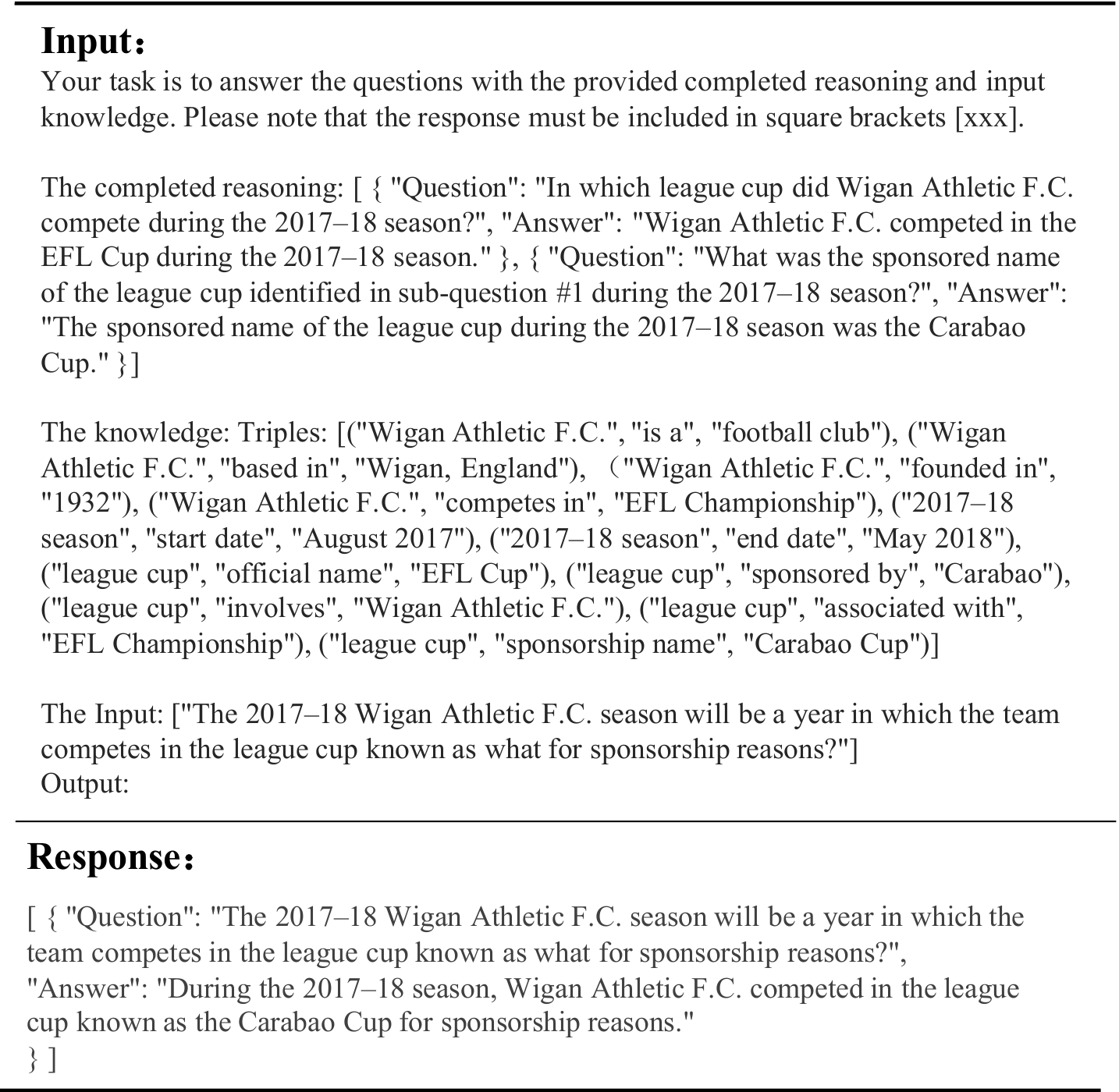}
  \caption{The prompt case of reasoning with the completed reasoning.}
  \label{fg:pcase3}
\end{figure*}

\begin{figure*}[h]
  \centering  \includegraphics[width=0.9\linewidth]{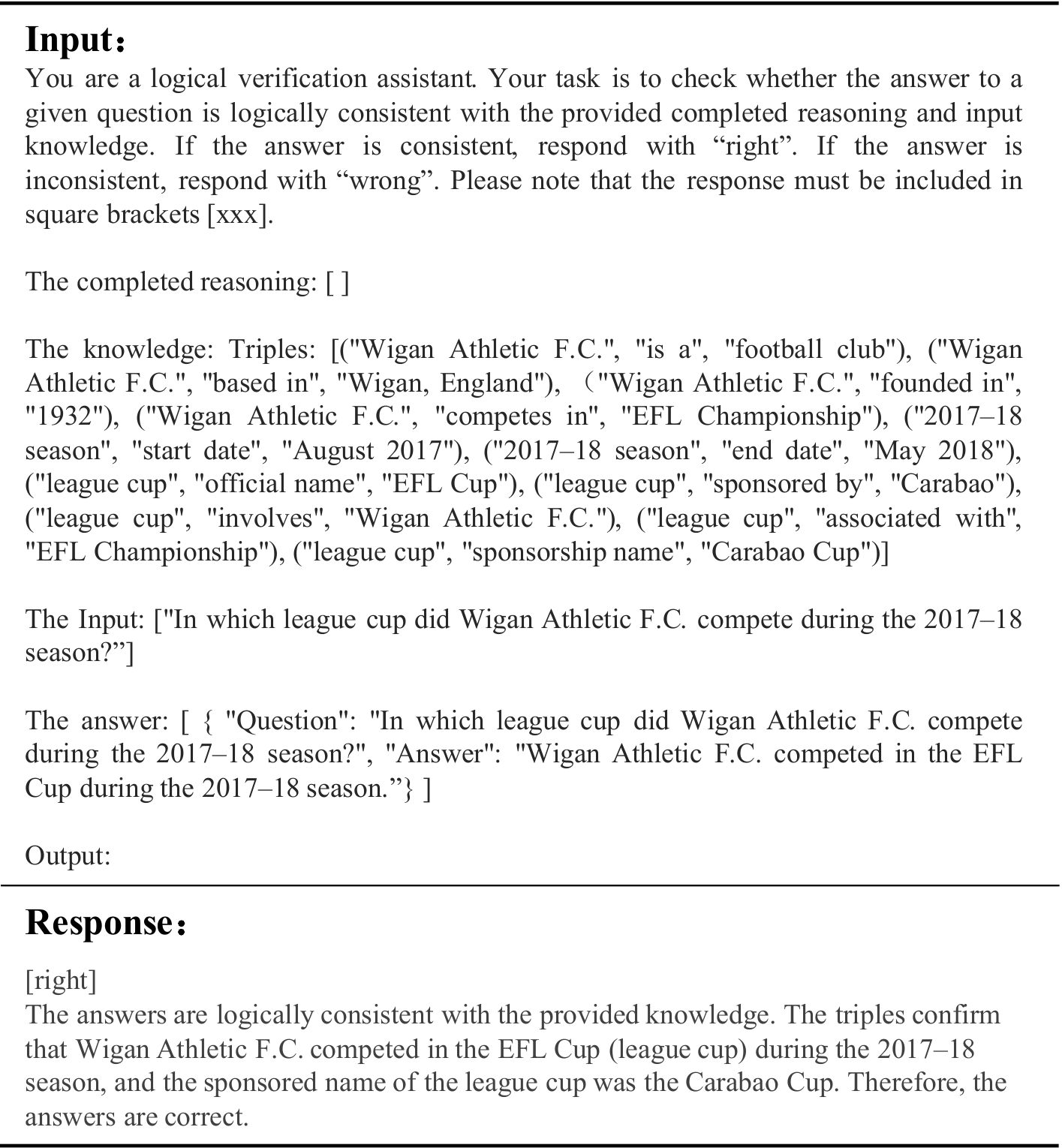}
  \caption{The prompt case of self-verification.}
  \label{fg:pcase2}
\end{figure*}

Additionally, Figure~\ref{fg:pcase1}, Figure~\ref{fg:pcase2} and Figure~\ref{fg:pcase3}  illustrate the process of prompting the large language model (LLM) to perform reasoning and self-verification, which provides a detailed breakdown of how the model generates and validates intermediate reasoning steps, ensuring the accuracy and reliability of the final output.

\section{Prompts in CogGRAG}
\label{sec:prompt}
In this section, we show all the prompts that need to be used in the main experiments as shown in Table~\ref{p:des}, Table~\ref{p:ext}, Table~\ref{p:ext_g}, Table~\ref{p:res}, Table~\ref{p:ver} and Table~\ref{p:rethink}.

\begin{table*}[h]
\centering
\resizebox{\textwidth}{!}{
\begin{tabular}{@{}p{\textwidth}@{}}
\toprule
\textbf{$\texttt{prompt}_{dec}$} \\
\midrule

\textbf{Prompt head:} ``Your task is to decompose the given question Q into sub-questions. You should based on the specific logic of the question to determine the number of sub-questions and output them sequentially.''\\
\\
\textbf{Instruction:} ``Please only output the decomposed sub-questions as a string in list format, where each element represents the text of a sub-question, in the form of \{[``subq1'', ``subq2'', ``subq3'']\}. For each sub-question, if you consider the sub-question to be sufficiently simple and no further decomposition is needed, then output ``End.'', otherwise, output ``Continue.'' Please strictly follow the format of the example below when answering the question.
\\
Here are some examples:''\\

``Input: ``What year did Guns N Roses perform a promo for a movie starring Arnold Schwarzenegger as a former New York Police detective?''\\
Output: [
    \{
        ``Sub-question'': ``What movie starring Arnold Schwarzenegger as a former New York Police detective is being referred to?'',
        ``State'': ``Continue.''
    \},
    \{
        ``Sub-question'': ``In what year did Guns N Roses perform a promo for the movie mentioned in sub-question \#1?'',
        ``State'': ``End.''
    \}
]''\\

``Input: ``What is the name of the fight song of the university whose main campus is in Lawrence, Kansas and whose branch campuses are in the Kansas City metropolitan area?''\\
Output: [
    \{
        ``Sub-question'': ``Which university has its main campus in Lawrence, Kansas and branch campuses in the Kansas City metropolitan area?'',
        ``State'': ``End.''
    \},
    \{
        ``Sub-question'': ``What is the name of the fight song of the university identified in sub-question \#1?'',
        ``State'': ``End.''
    \}
]''\\

``Input: ``Are the Laleli Mosque and Esma Sultan Mansion located in the same neighborhood?''\\
Output: [
    \{
        ``Sub-question'': ``Where is the Laleli Mosque located?'',
        ``State'': ``End.''
    \},
    \{
        ``Sub-question'': ``Where is the Esma Sultan Mansion located?'',
        ``State'': ``End.''
    \},
    \{
        ``Sub-question'': ``Are the locations of the Laleli Mosque and the Esma Sultan Mansion in the same neighborhood?'',
        ``State'': ``End.''
    \}
]''\\
\\
``\textbf{Input:} \textcolor{blue}{Question $Q$}''\\
``\textbf{Output:} '' \\
\bottomrule
\end{tabular}
}
\caption{The prompt template for decomposition.}
\label{p:des}
\end{table*}

\begin{table*}[h]
\centering
\resizebox{\textwidth}{!}{
\begin{tabular}{@{}p{\textwidth}@{}}  % 段落型列，支持自动换行
\toprule
\textbf{$\texttt{prompt}_{ext}$} \\
\midrule
\textbf{Prompt head:} ``Your task is to extract the entities (such as people, places, organizations, etc.) and relations (representing behaviors or properties between entities, such as verbs, attributes, or categories, etc.) involved in the input questions. These entities and relations can help answer the input questions.''\\
\\
\textbf{Instruction:} ``Please extract entities and relations in one of the following forms: entity, tuples, or triples from the given input List. Entity means that only an entity, i.e. <entity>. Tuples means that an entity and a relation, i.e. <entity-relation>. Triples means that complete triples, i.e. <entity-relation-entity>. Please strictly follow the format of the example below when answering the question.''\\
\\
``\textbf{Input:} \textcolor{blue}{[The mind map $\mathcal{M}$]}''\\
``\textbf{Output:}''\\
\bottomrule
\end{tabular}
}
\caption{The prompt template for extraction on local level.}
\label{p:ext}
\end{table*}

\begin{table*}[h]
\centering
\resizebox{\textwidth}{!}{
\begin{tabular}{@{}p{\textwidth}@{}}
\toprule
\textbf{$\texttt{prompt}_{ext}$} \\
\midrule
\textbf{Prompt head:} ``Your task is to extract the subgraphs involved in a set of input questions.''\\
\\
\textbf{Instruction:} ``Please extract and organize information from a set of input questions into structured subgraphs. Each subgraph represents a group of triples (subject, relation, object) that share common entities and capture the logical relationships between the questions. Here are some examples:''\\

``Input: [``What is the capital of France?'', ``Who is the president of France?'', ``What is the population of Paris?'']\\
Output: [(``France'', ``capital'', ``Paris''), (``France'', ``president'', ``Current President''), (``Paris'', ``population'', ``Population Number'')]''\\
\\
``\textbf{Input:} \textcolor{blue}{[The mind map $\mathcal{M}$]}''\\
``\textbf{Output:}'' \\
\bottomrule
\end{tabular}
}
\caption{The prompt template for extraction on global level.}
\label{p:ext_g}
\end{table*}

\begin{table*}[h]
\centering
\resizebox{\textwidth}{!}{
\begin{tabular}{@{}p{\textwidth}@{}}
\toprule
\textbf{$\texttt{prompt}_{res}$} \\
\midrule
\textbf{Prompt head:} ``Your task is to answer the questions with the provided completed reasoning and input knowledge.''\\
\\
\textbf{Instruction:} ``Please note that the response must be included in square brackets [xxx].''\\
\\
``\textbf{The completed reasoning:} \textcolor{blue}{[The set of verified answers $\hat{\mathcal{M}}$]}''\\
\\
``\textbf{The knowledge graph:} \textcolor{blue}{[The final retrieved triple set $\mathcal{T}$]}''\\
\\
``\textbf{Input:} \textcolor{blue}{[Subquestion $q^t$]}''\\

``\textbf{Output:}''\\
\bottomrule
\end{tabular}
}
\caption{The prompt template for reasoning.}
\label{p:res}
\end{table*}

\begin{table*}[h]
\centering
\resizebox{\textwidth}{!}{
\begin{tabular}{@{}p{\textwidth}@{}}
\toprule
\textbf{$\texttt{prompt}_{ver}$} \\
\midrule
\textbf{Prompt head:} ``You are a logical verification assistant. Your task is to check whether the answer to a given question is logically consistent with the provided completed reasoning and input knowledge. If the answer is consistent, respond with ``right''. If the answer is inconsistent, respond with ``wrong''.''\\
\\
\textbf{Instruction:} ``Please note that the response must be included in square brackets [xxx].''\\
\\
``\textbf{The completed reasoning:} \textcolor{blue}{[The set of verified answers $\hat{\mathcal{M}}$]}''\\
\\
``\textbf{The knowledge graph:} \textcolor{blue}{[The final retrieved triple set $\mathcal{T}$]}''\\
\\

``\textbf{Answer:} \textcolor{blue}{[The candidate answer $a^{t}$]}''\\
\\

``\textbf{Input:} \textcolor{blue}{[Subquestion $q^t$]}''\\

``\textbf{Output:}''\\
\bottomrule
\end{tabular}
}
\caption{The prompt template for self-verification.}
\label{p:ver}
\end{table*}

\begin{table*}[h]
\centering
\resizebox{\textwidth}{!}{
\begin{tabular}{@{}p{\textwidth}@{}}
\toprule
\textbf{$\texttt{prompt}_{rethink}$} \\
\midrule
\textbf{Prompt head:} ``You are a reasoning and knowledge integration assistant. Your task is to re-think a question that was previously answered incorrectly by the self-verification model. Use the provided completed reasoning and input knowledge to generate a new answer.''\\
\\
\textbf{Instruction:} ``Please note, if the knowledge is insufficient to answer the question, respond with ``Insufficient information, I don't know''. The response must be included in square brackets [xxx].''\\
\\
``\textbf{The completed reasoning:} \textcolor{blue}{[The set of verified answers $\hat{\mathcal{M}}$]}''\\
\\
``\textbf{The knowledge graph:} \textcolor{blue}{[The final retrieved triple set $\mathcal{T}$]}''\\
\\
``\textbf{Input:} \textcolor{blue}{[Subquestion $q^{t}$]}''\\

``\textbf{Output:}''\\
\bottomrule
\end{tabular}
}
\caption{The prompt template for re-thinking.}
\label{p:rethink}
\end{table*}

\end{document}